%% file: acl_latex.tex
\pdfoutput=1
\documentclass[11pt]{article}

\usepackage[]{acl}
\usepackage{times}
\usepackage{latexsym}
\usepackage[T1]{fontenc}
\usepackage[utf8]{inputenc}
\usepackage{microtype}
\usepackage{listings}
\usepackage{graphicx}
\usepackage{booktabs}
\usepackage{multirow}
\usepackage{multicol}
\usepackage{arydshln}
\usepackage{placeins}
\usepackage{stfloats}
\usepackage{subcaption}
\usepackage{tikz}
\usepackage{pgfplots}
\pgfplotsset{compat=1.17}
\usepackage[hang,flushmargin]{footmisc}

\newcommand\unnumberedfootnote[1]{%
\begingroup\renewcommand\thefootnote{}\footnote{#1}\addtocounter{footnote}{-1}%
\endgroup
}

\title{Exploring Data Augmentation for Code Generation Tasks}

\author{
  Pinzhen Chen\textsuperscript{*} \\
  School of Informatics \\
  University of Edinburgh \\
  \texttt{pinzhen.chen@ed.ac.uk} \\
  \And
  Gerasimos Lampouras \\
  Noah's Ark Lab \\
  Huawei \\
  \texttt{gerasimos.lampouras@huawei.com}\\
  }

\begin{document}
\maketitle
\begin{abstract}
Advances in natural language processing, such as transfer learning from pre-trained language models, have impacted how models are trained for programming language tasks too. Previous research primarily explored code pre-training and expanded it through multi-modality and multi-tasking, yet the data for downstream tasks remain modest in size. Focusing on data utilization for downstream tasks, we propose and adapt augmentation methods that yield consistent improvements in code translation and summarization by up to 6.9\% and 7.5\% respectively. Further analysis suggests that our methods work orthogonally and show benefits in output code style and numeric consistency. We also discuss test data imperfections.\unnumberedfootnote{\textsuperscript{*}Work done during an internship at Huawei Noah's Ark Lab. Our code will be available at \url{https://github.com/huawei-noah/noah-research/tree/master/NLP/DA4CodeGeneration}}
\end{abstract}

\section{Introduction}
Recent years have seen the rapid development of pre-trained models (PLMs) to enable knowledge transfer from generic texts to specific downstream tasks \citep{devlin-etal-2019-bert, liu-etal-2019-roberta}. PLMs have been applied to the programming language domain as well, following the same paradigm of (continuing) training PLMs on code and text data, and then fine-tuning them for specific tasks \citep{kanade-etal-2020-learning, feng-etal-2020-codebert}. PLMs are often adapted to programming languages by including code-specific modalities as part of the input like serialized syntax trees and data flows \citep{guo-etal-2021-graphcodebert,guo-etal-2022-unixcoder,tipirneni-etal-2022-structcoder}. Such works have outperformed rule-based tools in various tasks, e.g. the CodeXGLUE benchmark \citep{lu-etal-2021-codexglue}.

Despite the abundance of raw code available for pre-training, code data that meet downstream needs stay modest in size. This is due to the fact that, unlike texts, code datasets cannot be easily curated by people without programming knowledge. For example, code translation data in CodeXGLUE is sized at 10K, which is orders of magnitude smaller than their natural language counterparts that often include millions of instances \citep{kocmi-etal-2022-findings}.

We are therefore motivated to enrich data in the fine-tuning phase of code PLMs, using automatic data augmentation (DA) methods like back-translation, monolingual, multilingual, and numeric augmentation. We extensively experiment on code translation, where a programming language is converted to another, and summarization, where a textual description is produced from a code block. Even with limited resources, we can lift performance by 6.9\% for translation and 7.5\% for summarization compared to baselines. Through manual inspection and extra evaluation measures, we demonstrate that our methods lead to desirable enhancements special to code, namely better output code style and number correctness.

\section{Methodology}
\subsection{Data synthesis}
Back-translation \citep[BT,][]{sennrich-etal-2016-improving} is a data augmentation technique originated from machine translation, where an auxiliary model is used to construct pseudo-parallel data from monolingual resources. It can be straightforwardly applied to code translation. Formally, to train a model $f()$ that converts a programming language $PL_{x}$ into $PL_{y}$, we first train an inverse model $g(PL_{y})\rightarrow PL_{x}$ with the same parallel data. Having the inverse model $g()$, extra monolingual data in $PL_{y}$ is translated into $PL_{x}^\prime$ to form pseudo-parallel pairs $PL_{x}^\prime$-$PL_{y}$ that can be used to train $f()$.

For code summarization, back-translation is not applicable as ``monolingual'' natural language ($NL$) summaries unaligned to code hardly exist. Hence we propose to use the summaries originally associated with a single programming language as a pivot for other programming languages. After inversing code-to-text data which has source side code available in multiple programming languages ($PL_{1}\rightarrow NL,\ldots,PL_{n}\rightarrow NL$), we train a multilingual text-to-code generator, which outputs a designated programming language given a natural language summary and a target language tag ($NL+ tag_{\{1,\ldots n\}}\rightarrow\{PL_{1},\ldots,PL_{n}$\}). This generator can iteratively produce code in different $PL$s by inputting summaries regardless of the original $PL\rightarrow NL$ alignment. These synthesized data, despite having a lower quality, can augment the training data for summarization.

\subsection{Utilization of multilinguality}
\citet{currey-etal-2017-copied} suggested that including monolingual data in the target language as an additional autoencoding (AE) objective benefits translation models trained on limited data. We migrate this objective to code translation by mixing $PL_{x}\rightarrow PL_{y}$ and $ PL_{y}\rightarrow PL_{y}$ data. This effectively builds a multilingual encoder that enables knowledge transfer, given the high similarity between programming languages, namely the overlap of numerals, syntax tokens, reserved keywords, etc. This process constrains the decoder side to a single programming language $PL_{y}$ to not add complexity.

In code summarization, as the target NL is fundamentally divergent from the input PL, the autoencoding objective might not be useful. In contrast, we train a ``multilingual'' code summarization model $\{PL_{1}, \ldots, PL_{n}\}\rightarrow NL$ where the system takes an arbitrary programming language to produce a natural language summary. Such a many-to-one model allows encoder knowledge sharing too and exposes the decoder to more NL summaries.

\subsection{Numeric awareness}
Referenced variables and their values are unique components of programming languages; to enhance understanding of these values, previous works on pre-training suggested attending to appropriate modalities, e.g. data flow \citep{guo-etal-2021-graphcodebert}. Such sophisticated handling of values might not be necessary for code translation, as copying them over to the target suffices. However, given a small training size, any translation model will still only be exposed to sparse numerical input. To increase model robustness, we augment the data by creating new instances where, in all code tokens containing a number, each digit is randomly replaced with another digit, consistently on both the source and target sides. We do not distinguish between purely numerical tokens and tokens including a number. For instance, a variable ``num1'' could become ``num4'' in the augmented code pair. The method guarantees that the number-swapped synthetic code is grammatical and compilable.

\begin{figure}
    \centering
    \includegraphics[scale=0.73,trim={4pt 0pt 4pt 0pt},clip]{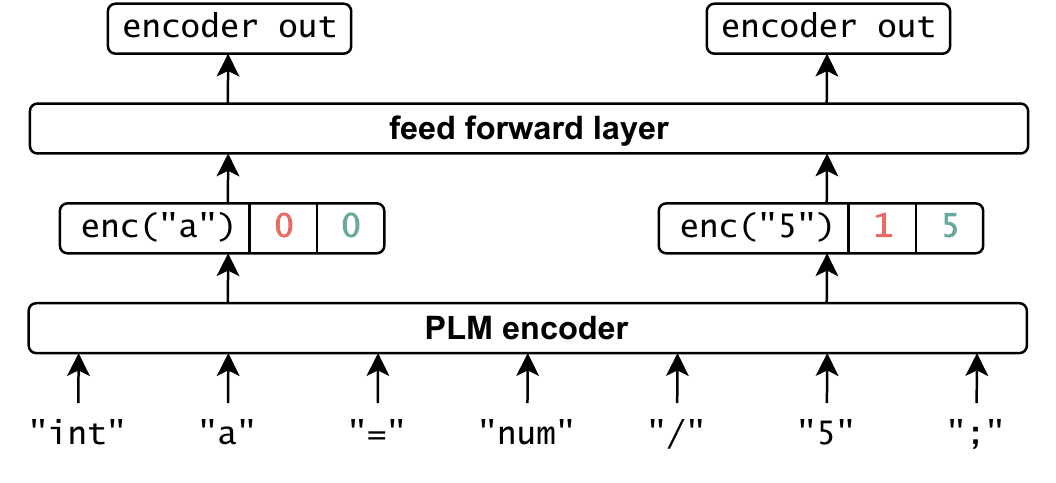}
    \caption{Numeric encoding with a PLM encoder, exemplifying how ``a'' and ``5'' are encoded differently.}
    \label{fig:number-model}
\end{figure}

Apart from numerical augmentation, we propose to include input numbers directly in the encoder output as mathematical values, complementary to their string embedding representations. As illustrated in Figure~\ref{fig:number-model}, we append two dimensions to the original encoder output. Particularly, one dimension (red, left) is a binary value (0/1) indicating whether the respective input is a number, while the other dimension (green, right) inherits the input's value, or 0 if the input is not numeric. The expanded embedding can be reduced to its original size via a feed-forward layer; such a change requires no modification to the pre-trained encoder.

\section{Experiments}
\subsection{Tasks, datasets and evaluation}
We benchmark our methods on the code task suite CodeXGLUE \citep{lu-etal-2021-codexglue}. Its translation task uses code originally developed in Java and then migrated to C\#, so the corresponding C\#-Java snippets are considered parallel. Training, validation, and test sizes are 10K, 0.5K, and 1K. For back-translation, we translated 377K lines of monolingual Java, albeit out-of-domain, from other CodeXGLUE tasks, into C\#. To ensure that the target side consists of genuine data, we only experimented with the C\#$\rightarrow$Java direction as there is no other C\# code in the benchmark for BT.

The summarization task employs CodeSearchNet \citep{husain-etal-2019-codesearchnet} and covers six languages: Ruby, JavaScript, Go, Python, Java, and PHP. Training sizes range from 25K to 250K, totalling 908K; validation and test sets are between 1K and 15K. We performed multilingual back-translation by reversing the dataset so no external data is introduced; this leads to a five-fold BT data of $4.5$M (908K$\times$5). All programming languages share an equal amount of original and synthetic data combined. Moreover, to compare the quality of neural back-translation against hand-written rule-based conversion, we created 80K JavaScript-summary pairs from Python-summary data using \texttt{jsbuilder}.

We report code translation results in BLEU-4 \citep{papineni-etal-2002-bleu}, exact line matches (EM, in \%), and CodeBLEU, a weighted sum of four accuracies: n-grams, weighted n-grams, syntax, and data flow \citep{ren-etal-2020-codebleu}. Code summarization performance is measured by the de facto choice of BLEU-4 on natural language texts.

\subsection{Systems}
For all tasks, we use the CodeXGLUE baseline, i.e. CodeBERT with a Transformer decoder, for our base and inverse models (for data synthesis). We continue training PLMs on the augmented data, then fine-tune on the original data, except for numeric augmentation where we mix the synthetic data with the training set. Monolingual and multilingual summarization experiments share the same configurations. For numeric encoding with CodeBERT, we add a feed-forward layer to make the baseline as deep as our proposed network.

To provide results with stronger baselines, we also test with GraphCodeBERT \citep{guo-etal-2021-graphcodebert} for translation and UniXcoder \citep{guo-etal-2022-unixcoder} for summarization. This helps to verify the stability of data augmentation performance across distinct PLM architectures. We stick to the relevant PLMs' hyperparameters except for batch size. Model and training details, with links to the preprocessing and evaluation scripts, can be found in Appendix~\ref{sec:appendix-config}.

\begin{table}[tbp]
\input{tabular_translation_results.tex}
\caption{Test results for C\#$\rightarrow$Java translation.}
\label{tab:translation-results}
\end{table}

\begin{table}[tbp]
\input{tabular_summarization_results.tex}
\caption{Test results for code summarization in BLEU.}
\label{tab:summarization-results}
\end{table}

\begin{table*}[tbp]
\input{tabular_numeric_augmentation.tex}
\caption{Test results for C\#$\rightarrow$Java translation with numeric augmentation and encoding.}
\label{tab:numeric-augmentation}
\end{table*}

\subsection{Results and Discussions}
We first show translation results in Table~\ref{tab:translation-results}, where back-translation surpasses baselines by a large margin for both PLMs; on top of it, autoencoding brings a small gain. Table~\ref{tab:summarization-results} indicates that back-translation also steadily helps code summarization overall. An interesting pattern from both PLMs is that BT helps Ruby and Java less than other languages. Furthermore, learning a single multilingual model is better than learning separate monolingual models, potentially due to transfer learning between programming languages and the increase in natural language data size on the output side.

\begin{table*}[tbp]
\input{tabular_two_augmentation.tex}
\caption{Test results for C\#$\rightarrow$Java translation with multiple augmentation techniques.}
\label{tab:two-augmentation}
\end{table*}

Table~\ref{tab:numeric-augmentation} reports the results for numeric augmentation and numeric encoding in translation. Adding number-swapped data to training surpasses the baseline, while our numeric encoding proposal under-performs the baseline. To accommodate the neural network weights which are orders of magnitude smaller than the variable values encountered in code, we investigate linear and logarithmic value scaling. As the scaling gets smaller, result numbers gradually catch up; the optimal is a logarithmic transformation, whereby the model attains the highest performance.

To directly assess our value-aware augmentation, we compute and append output token accuracies to Table~\ref{tab:numeric-augmentation}, with a distinction between numeric and non-numeric tokens. We can observe that the numerical approaches aid number generation without compromising non-numbers, and the improvement in number correctness is generally consistent with the improvement in BLEU and EM. Additional visualization in Appendix~\ref{sec:appendix-code-number} implies that DA models can maintain numeric consistency even when the output is extremely long and complicated.\label{sec:number-consistency}

Finally, Table~\ref{tab:two-augmentation} examines if the above methods, namely back-translation and numeric augmentation, work orthogonally. It is observed that better results are achieved when numeric augmentation is applied to the original data, but not to the back-translated data. This is probably because BT is already of inferior quality, so numerical augmentation introduces extra noise. Nevertheless, combining BT and AE with numeric augmentation over the original data leads to the best outcome.

\section{Analysis}
Upon inspecting the translation test outputs, we find that our data-augmented model is better exposed to the target Java language: it has learned the Java programming conventions instead of following the input code style. We present test instances focused on element retrieval methods, by listing sources, references, and outputs from the CodeBERT baseline and our BT-augmented model in Table~\ref{code:analysis}. Whilst direct retrieval of an element through reference to its position is possible in Java, we observe that the baseline tends to imitate the code style in source C\#, but the DA model closely follows the Java coding convention where the inbuilt method \texttt{get()} is favoured over directly accessing the attributes by indices.

\begin{table*}[t!]
\centering
\input{analysis-code-translation-outputs.tex}
\caption{C\#-Java output translations of element retrieval methods, before and after data augmentation.}
\label{code:analysis}
\end{table*}

We should note that in the translation test set a small proportion of code pairs seem to be divergent, which can lead to an inaccurate estimate of translation performance. We record a few examples of these imperfections in Appendix~\ref{sec:appendix-test-inspection}, but leave in-depth investigation and refinement for future work.\label{sec:test-inspection}

\section{Related Works}
Recent research at the intersection of natural language processing and programming languages concentrated on pre-training. \citet{kanade-etal-2020-learning} trained CuBERT to obtain embeddings for code understanding tasks. \citet{feng-etal-2020-codebert} developed CodeBERT by training RoBERTa on bimodal text-code data with replaced token detection \citep{clark-etal-2020-electra}. In GraphCodeBERT, \citet{guo-etal-2021-graphcodebert} incorporated data flow edge prediction and data-variable alignment. Researchers expanded decoder-only models to the code domain too, e.g. CodeGPT, Codex, and Pangu-Coder \citep{lu-etal-2021-codexglue,chen-etal-2021-evaluating-large,christopoulou-etal-2022-pangu}. Universal encoder-decoder code PLMs have also been presented: PyMT5, CodeT5, PLBART, UniXcoder, and StructCoder \cite{clement-etal-2020-pymt5,wang-etal-2021-codet5,ahmad-etal-2021-unified,guo-etal-2022-unixcoder,tipirneni-etal-2022-structcoder}. UniXcoder, which we used, adopts attention masks to control encoder-decoder behaviours in a shared encoder-decoder network.

Datasets for specific tasks concerning code are usually small, so data augmentation can help to boost performance. \citet{roziere-etal-2020-unsupervised} combined cross-lingual masked modelling and iterative back-translation to build an unsupervised code transcompiler. \citet{ahmad-etal-2022-summarize} ran code-to-text summarization then text-to-code generation, to obtain translation data. In contrast, we train a text-to-code generation model by reversing the summarization data; our methods differ in both the procedure and the intended task. Also, \citet{yu-etal-2022-data} crafted rules for source code transformation, whilst our investigation is on automatic neural methods. Finally, techniques like dead code insertion and variable renaming in malware obfuscation \cite{you-yim-2010-malware}, as well as string manipulation (e.g. token noising, swapping, deletion) can be useful. Nonetheless, these methods are not task-specific, meaning they could be more appropriate for the generic code pre-training stage.

\section{Conclusion}
We adapt several data augmentation techniques to programming language translation and summarization. Our investigation includes data synthesis, knowledge sharing via multilinguality, and numeric-aware techniques. Enhanced performance is observed in experiments conducted on a variety of pre-trained code language models, and our analysis demonstrates that these methods can benefit output code style and numeric correctness.

\section{Limitations}
We identify the main limitation to lie in evaluation since we relied on automatic text metrics for both code and text generation. Ideally, code should be treated with software testing practices such as code review, compilation, unit testing, etc. Evaluation is further undermined given the test data issues revealed in Section~\ref{sec:test-inspection} and Appendix~\ref{sec:appendix-test-inspection}, so more human analysis should be of interest.

We also do not cover all potential code generation tasks, e.g. code synthesis, where a code snippet is created given a textual description. In this task, the source side carries much less information than the target. We apply a back-translation-style augmentation, but it does not significantly surpass the state-of-the-art PLM. Due to space constraints, we offer some preliminary views in Appendix~\ref{sec:appendix-code-synthesis}.

\section*{Acknowledgements}
We are grateful to Ignacio Iacobacci for his comments on numeric input scaling, and to the reviewers for their suggestions on qualitative analysis. 
We also thank the MindSpore team for providing technical support.\footnote{\url{https://www.mindspore.cn/en}}\textsuperscript{,}\footnote{\url{https://github.com/mindspore-ai}}

Pinzhen Chen is supported by UK Research and Innovation under the UK government’s Horizon Europe funding guarantee [grant number 10052546 -- High Performance Language Technologies].

\bibliography{custom, anthology}
\bibliographystyle{acl_natbib}

\appendix

\begin{table*}[b]
\input{tabular_model_configs.tex}
\caption{Model and training configurations.}
\label{tab:config}
\end{table*}

\begin{table*}[htb]
\centering
\input{analysis-appendix-code-imperfections.tex}
\caption{C\#-Java test instances that are not perfectly parallel, with divergence shown in bold.}
\label{code:test}
\end{table*}

\section{Model Configurations}
\label{sec:appendix-config}
Our training and model configurations are summarized here and in Table~\ref{tab:config}. We retain the relevant PLMs' default configurations as much as possible, except for a grid search on the learning rate for code summarization with UniXcoder. We also changed the batch size to utilize our GPUs.

The randomly initialized Transformer decoder attached to CodeBERT and GraphCodeBERT has 6 layers, 12 heads, 768 hidden dimensions, and other hyperparameters as default in PyTorch. For the numeric encoding experiments with CodeBERT, we append 2 dimensions to CodeBERT's 768d encoder output, then transform it back to 768d using a linear layer. To ensure a fair comparison, a 768d-to-768d layer is added to the baseline to make it as deep. 

All experiments are given a fixed budget to run. We save the best checkpoint according to validation BLEU. Results in the paper are based on a single run, but the experiments were benchmarked on PLMs of different architectures to reflect stability.

\section{More Inspections on Translation Test}
\subsection{Test imperfections}
\label{sec:appendix-test-inspection}
We show a few translation test instances that are not perfectly parallel in Table~\ref{code:test}. In these cases, the code in two languages will not function exactly the same when being executed.

\subsection{Numeric consistency}
\label{sec:appendix-code-number}
Complementing the number accuracy figures reported in Section~\ref{sec:number-consistency}, we list translation outputs containing numbers in Table~\ref{code:number} for visualization. It conveys the idea that our DA models can ensure number consistency even in very long and complicated outputs. In the baseline outputs, for example in test \#436, number incorrectness further leads to undesirable hallucinations, which can be prevented in the DA model's output.

\section{Code Synthesis with Augmentation}
\label{sec:appendix-code-synthesis}
For code synthesis, while reversing the summarization data is a natural solution, the difficulty lies in forming the class environment (visible and usable variables and methods) because. We parse the code in a summarization instance to obtain positive tokens, as well as randomly sample tokens from other genuine code as negative signals. In other words, from $PL\rightarrow NL$ pairs, we construct code synthesis data $NL+parse(PL)+random(PL^\prime)\rightarrow PL$.

We experiment on CodeXGLUE's code synthesis task, which samples data from CONCODE \citep{iyer-etal-2018-mapping} at 100K/2K/2K for training/validation/test. The source contains a text description as well as the available class variable and function names, and the target is the corresponding Java code. We reverse the Java summary data to create 181K synthetic data; to get available variable and method names, the code is parsed by \texttt{javalang} into tokens. Following CodeXGLUE, we use CodeGPT-adapted as a base model; we further experiment with StructCoder \citep{tipirneni-etal-2022-structcoder} which is a more up-to-date code PLM.

The outputs are evaluated by BLEU, EM, and CodeBLEU, similar to translation. Note that the test references are not publicly available, and test predictions need to be sent to the CodeXGLUE authors for evaluation, so we report results on both the validation and test set for reproducibility.

We notice that for CodeGPT, our augmentation work yields a small gain on validation and test sets. However, it does not improve upon the latest PLM for a few possible reasons: 1) StructCoder is remarkably stronger than CodeGPT, thus the room for improvement is small; 2) the summarization data we used to augment the synthesis task could be different in terms of topic, length, style, etc, resulting in a domain drift.

\begin{table}[htbp]
\input{tabular_synthesis_results.tex}
\caption{Results for code synthesis.}
\label{tab:code-synthesis}
\end{table}

\begin{table*}[htb]
\centering
\input{analysis-appendix-code-number.tex}
\caption{C\#-Java output translations containing numbers, before and after data augmentation.}
\label{code:number}
\end{table*}

\end{document}

%% file: tabular_translation_results.tex
\centering
\small
\setlength{\tabcolsep}{0.5ex}
\begin{tabular}{lccc}
\toprule
    & \textbf{BLEU} & \textbf{EM} & \textbf{CodeBLEU}\textsuperscript{$\dagger$} \\
\midrule
    \multicolumn{4}{l}{\textit{CodeBERT}} \\
    \hspace{0.5ex}paper     & 72.14 & 58.0 & - \\
    \hspace{0.5ex}replicate & 72.92 & 57.4 & 78.93\hspace{0.5ex}(72.92\hspace{0.5ex}/\hspace{0.5ex}73.61\hspace{0.5ex}/\hspace{0.5ex}87.08\hspace{0.5ex}/\hspace{0.5ex}82.10) \\
    \hspace{0.5ex}BT          & 77.34 & 61.4 & 83.36\hspace{0.5ex}(77.34\hspace{0.5ex}/\hspace{0.5ex}78.11\hspace{0.5ex}/\hspace{0.5ex}\textbf{90.34}\hspace{0.5ex}/\hspace{0.5ex}87.64) \\
    \hspace{0.5ex}\hspace{0.5ex}+ AE     & \textbf{77.60} & \textbf{61.8} & \textbf{83.47}\hspace{0.5ex}(\textbf{77.60}\hspace{0.5ex}/\hspace{0.5ex}\textbf{78.30}\hspace{0.5ex}/\hspace{0.5ex}90.02\hspace{0.5ex}/\hspace{0.5ex}\textbf{87.96}) \\
\midrule
    \multicolumn{4}{l}{\textit{GraphCodeBERT}} \\
    \hspace{0.5ex}paper & 72.64 & 58.8 & - \\
    \hspace{0.5ex}replicate & 72.66 & 58.9 & 78.55\hspace{0.5ex}(72.66\hspace{0.5ex}/\hspace{0.5ex}73.35\hspace{0.5ex}/\hspace{0.5ex}87.44\hspace{0.5ex}/\hspace{0.5ex}80.74) \\
    \hspace{0.5ex}BT    & 75.15 & 60.7 & 82.13\hspace{0.5ex}(75.15\hspace{0.5ex}/\hspace{0.5ex}75.86\hspace{0.5ex}/\hspace{0.5ex}90.06\hspace{0.5ex}/\hspace{0.5ex}87.46) \\
    \hspace{0.5ex}\hspace{0.5ex}+ AE      & \textbf{76.15} & \textbf{62.5} & \textbf{82.88}\hspace{0.5ex}(\textbf{76.15}\hspace{0.5ex}/\hspace{0.5ex}\textbf{76.87}\hspace{0.5ex}/\hspace{0.5ex}\textbf{90.54}\hspace{0.5ex}/\hspace{0.5ex}\textbf{87.95}) \\
\bottomrule
\multicolumn{4}{l}{\footnotesize{\textsuperscript{$\dagger$}average (n-gram\hspace{0.5ex}/\hspace{0.5ex}weighted n-gram\hspace{0.5ex}/\hspace{0.5ex}syntax\hspace{0.5ex}/\hspace{0.5ex}data flow)}} \\
\end{tabular}

%% file: tabular_summarization_results.tex
\centering
\small
\setlength{\tabcolsep}{0.46ex}
\begin{tabular}{lccccccc}
\toprule
    & \textbf{Ruby} & \textbf{JS} & \textbf{Go} & \textbf{Py} & \textbf{Java} & \textbf{PHP} & \textbf{Avg.} \\
\midrule
    \textit{CodeBERT} \\
    \hspace{0.5ex}paper & 12.16 & 14.90 & 18.07 & 19.06 & 17.65 & 25.16 & 17.83 \\
    \hspace{0.5ex}monolingual & 12.39 & 14.13 & 17.89 & 18.22 & 18.66 & 25.14 & 17.73 \\
    \hspace{0.5ex}\hspace{0.5ex}+ rule-trans &   - & 15.35 &   - &   - &   - &   - & - \\
    \hspace{0.5ex}\hspace{0.5ex}+ BT & 13.76 & 15.00 & 18.30 & 18.60 & 19.64 & 25.69 & 18.50 \\
    \hspace{0.5ex}multilingual & \textbf{14.93} & 15.53 & 18.68 & 18.71 & 19.70 & 25.96 & 18.92 \\
    \hspace{0.5ex}\hspace{0.5ex}+ rule-trans & 14.58 & 15.65 & 18.77 & 18.95 & \textbf{19.86} & 25.98 & 18.97 \\
    \hspace{0.5ex}\hspace{0.5ex}+ BT & 14.91 & \textbf{15.81} & \textbf{18.88} & \textbf{18.97} & 19.69 & \textbf{26.10} & \textbf{19.06} \\
\midrule
    \textit{UniXcoder} \\
    \hspace{0.5ex}paper & 14.87 & 15.85 & 19.07 & 19.13 & 20.31 & 26.54 & 19.30 \\
    \hspace{0.5ex}monolingual & 14.81 & 15.28 & 18.93 & 19.05 & 20.22 & 26.66 & 19.16 \\
    \hspace{0.5ex}multilingual & \textbf{15.15} & 15.64 & 19.03 & 19.22 & \textbf{20.45} & 26.59 & 19.35 \\
    \hspace{0.5ex}\hspace{0.5ex}+ BT & 14.94 & \textbf{15.85} & \textbf{19.29} & \textbf{19.36} & 20.43 & \textbf{26.69} & \textbf{19.43} \\
\bottomrule
\end{tabular}

%% file: tabular_numeric_augmentation.tex
\centering\small
\begin{tabular}{lccccc}
\toprule
     & \multirow{2}{*}{\textbf{BLEU}} & \multirow{2}{*}{\textbf{EM}} & \multirow{2}{*}{\textbf{CodeBLEU}} & \multicolumn{2}{c}{\textbf{Token Accuracy}} \\
     & & & & \textbf{numeric} & \textbf{non-numeric} \\
\midrule
    \textit{CodeBERT} + FFN & 72.88 & 58.0 & 78.07 (72.88 / 73.66 / 86.15 / 79.59) & 74.50 & 86.72 \\
    \hspace{1ex}+ numeric augmentation & \textbf{74.00} & \textbf{59.5} & \textbf{79.43} (\textbf{74.00} / \textbf{74.72} / \textbf{87.01} / \textbf{82.00}) & \textbf{76.14} & \textbf{87.30} \\
\midrule
    numeric encoding & 72.95 & 58.1 & 78.77 (72.95 / 73.74 / 86.96 / 81.45) & 73.74 & 86.84 \\
    \multicolumn{6}{l}{\hspace{1ex}+ numeric augmentation with value scaling} \\
    \hspace{2ex}$\times 10^2$ & 71.32 & 51.6 & 77.71 (71.32 / 72.25 / 86.22 / 81.05) & 72.48 & 85.98 \\
    \hspace{2ex}$\times 1$ (no scaling) & 72.51 & 57.4 & 78.45 (72.51 / 73.38 / 86.47 / 81.46) & 72.92 & 86.49 \\
    \hspace{2ex}$\times 10^{-2}$ & 73.48 & \textbf{59.2} & 79.41 (73.48 / 74.28 / 87.31 / 82.56) & 74.11 & 87.11 \\
    \hspace{2ex}$\times 10^{-4}$ & 74.01 & 58.9 & 79.73 (74.07 / 74.75 / 87.29 / 82.87) & 74.93 & \textbf{87.48} \\
    \hspace{2ex}$\log _{10}()$ & \textbf{74.16} & 59.1 & \textbf{79.84} (\textbf{74.16} / \textbf{74.91} / \textbf{87.39} / \textbf{82.90}) & \textbf{75.22} & 87.32 \\
\bottomrule
\end{tabular}

%% file: tabular_two_augmentation.tex
\centering
\small
\begin{tabular}{lcccccc}
\toprule
     & \multirow{2}{*}{\textbf{BLEU}} & \multirow{2}{*}{\textbf{EM}} & \multirow{2}{*}{\textbf{CodeBLEU}} & \multicolumn{2}{c}{\textbf{Token Accuracy}} \\
     & & & & \textbf{numeric} & \textbf{non-numeric} \\
\midrule
    \textit{CodeBERT} replicate & 72.92 & 57.4 & 78.93 (72.92 / 73.61 / 87.08 / 82.10) & 74.64 & 87.54 \\
    BT & 77.34 & \textbf{61.4} & 83.36 (77.34 / 78.11 / 90.34 / 87.64) & 78.09 & 88.62 \\
    \hspace{1ex}+ num. aug. original only & \textbf{77.69} & 61.0 & \textbf{83.44} (\textbf{77.69} / \textbf{78.33} / 90.19 / 87.56) & \textbf{78.54} & \textbf{88.69} \\
    \hspace{1ex}+ num. aug. BT and original & 77.37 & 60.9 & 83.43 (77.37 / 78.07 / \textbf{90.36} / \textbf{87.94}) & 77.16 & 88.55 \\
    BT + AE & 77.60 & 61.8 & 83.47 (77.60 / 78.30 / 90.02 / \textbf{87.96}) & 77.16 & 88.64 \\
    \hspace{1ex}+ num. aug. original only & \textbf{77.96} & \textbf{62.0} & \textbf{83.63} (\textbf{77.96} / \textbf{78.62} / \textbf{90.15} / 87.82) & \textbf{78.01} & \textbf{88.79} \\
\bottomrule
\end{tabular}

%% file: analysis-code-translation-outputs.tex
\begin{lstlisting}[basicstyle=\linespread{0.8}\ttfamily\footnotesize,escapechar=@]
// @\textit{test \#85}@
@\textbf{C\# source}\hspace{7ex}@ ... GetEscherRecord(int index){return escherRecords@\textbf{[index]}@;}
@\textbf{Java reference}@ ... getEscherRecord(int index){return escherRecords@\textbf{.get(index)}@;}
@\textbf{baseline}\hspace{8.5ex}@ ... getEscherRecord(int index) {return escherRecords@\textbf{[index]}@;}
@\textbf{DA model}\hspace{8.5ex}@ ... getEscherRecord(int index) {return escherRecords@\textbf{.get(index)}@;}

// @\textit{test \#90}@
@\textbf{C\# source}\hspace{7ex}@ public virtual IQueryNode GetChild(){return GetChildren()@\textbf{[0]}@;}
@\textbf{Java reference}@ public QueryNode getChild() {return getChildren()@\textbf{.get(0)}@;}
@\textbf{baseline}\hspace{8.5ex}@ public QueryNode getChild() {return getChildren() @\textbf{== 0)}@;}
@\textbf{DA model}\hspace{8.5ex}@ public QueryNode getChild() {return getChildren()@\textbf{.get(0)}@;}

// @\textit{test \#978}@
@\textbf{C\# source}\hspace{7ex}@ public virtual SrndQuery GetSubQuery(int qn) { return m_queries@\textbf{[qn]}@; }
@\textbf{Java reference}@ public SrndQuery getSubQuery(int qn) {return queries@\textbf{.get(qn)}@;}
@\textbf{baseline}\hspace{8.5ex}@ public SrndQuery getSubQuery(int qn) {return queries@\textbf{[qn]}@;}
@\textbf{DA model}\hspace{8.5ex}@ public SrndQuery getSubQuery(int qn) {return queries@\textbf{.get(qn)}@; }

\end{lstlisting}

%% file: tabular_model_configs.tex
\centering\small
\begin{tabular}{ll} 
\toprule
\textbf{Hyperparameter} & \textbf{Value} \\
\midrule
PLM checkpoints & CodeBERT: \url{https://huggingface.co/microsoft/codebert-base} \\
 & GraphCodeBERT: \url{https://huggingface.co/microsoft/graphcodebert-base} \\
 & UniXcoder: \url{https://huggingface.co/microsoft/unixcoder-base} \\
 & CodeGPT: \url{https://huggingface.co/microsoft/CodeGPT-small-java-adaptedGPT2} \\
 & StructCoder: \url{https://github.com/reddy-lab-code-research/structcoder} \\
trainable parameters & CodeBERT: 172.5M \\
           & \hspace{1ex}+ numeric encoding: + 591k \\
           & GraphCodeBERT: 172.5M \\
           & UniXcoder: 126.5M \\
           & CodeGPT: 124.4M \\
           & StructCoder: 223.4M \\
learning rate & translation: 5e\textsuperscript{-5} \\
              & summarization: 1e\textsuperscript{-5}, 5e\textsuperscript{-5}, \textbf{1e\textsuperscript{-6}} , 5e\textsuperscript{-6} \\
              & synthesis: 5e\textsuperscript{-5} \\
optimizer & Adam (epsilon=1e\textsuperscript{-8})\\
training loss & cross-entropy (perplexity) \\
validation metric & best BLEU \\
beam size & 10 \\
CodeXGLUE & \url{https://github.com/microsoft/CodeXGLUE} \\
\texttt{jsbuilder} & \url{https://github.com/tvst/jsbuilder} \\
\texttt{javalang} & \url{https://github.com/c2nes/javalang} \\
\bottomrule
\end{tabular}

%% file: analysis-appendix-code-imperfections.tex
\begin{lstlisting}[basicstyle=\linespread{1}\ttfamily\footnotesize,escapechar=@]
// @\textit{test \#307}@
@\textbf{C\# source}\hspace{7ex}@ public override string ToString(){return "IndexSearcher("
@\hspace{23.5ex}@ + reader + "; executor=" + executor + ")";}
@\textbf{Java reference}@ public String toString() {return "IndexSearcher("
@\hspace{23.5ex}@ + reader + "; executor=" + executor
@\hspace{23.5ex}@ + "; @\textbf{sliceExecutionControlPlane " + sliceExecutor}@ + ")";}

// @\textit{test \#518}@
@\textbf{C\# source}\hspace{7ex}@ public override PushConnection OpenPush() throw 
@\hspace{23.5ex}@ {@\textbf{new NGit.Errors.NotSupportedException(}@ 
@\hspace{27.5ex}@ @\textbf{JGitText.Get().pushIsNotSupportedForBundleTransport}@);}
@\textbf{Java reference}@ public PushConnection openPush() throws 
@\hspace{23.5ex}@ {@\textbf{TransportException return new TcpPushConnection()}@;}

// @\textit{test \#892}@
@\textbf{C\# source}\hspace{7ex}@ public Builder()@\textbf{: base()\{lastDocID = -1;wordNum = -1;word = 0;\}}@
@\textbf{Java reference}@ public Builder() @\textbf{\{this(true);\}}@

// @\textit{test \#902}@
@\textbf{C\# source}\hspace{7ex}@ public override string ToString(){return "term="+ term+", field=" 
@\hspace{24.5ex}@ +field+", value="+value;}
@\textbf{Java reference}@ public String toString() {return "term="+term+", field=" 
@\hspace{24.5ex}@ +field+", value="+valueToString()@\textbf{+",docIDUpto="+docIDUpto}@;}
\end{lstlisting}

%% file: tabular_synthesis_results.tex
\centering
\small
\begin{tabular}{lccc}
\toprule
 & \textbf{BLEU} & \textbf{EM} & \textbf{CodeBLEU} \\
\midrule
    \multicolumn{4}{l}{\textit{CodeGPT} on validation} \\
    \hspace{1ex}replicate & 28.13 & 16.1 & 31.65 \\
    \hspace{1ex}\hspace{1ex}+ augmentation & \textbf{29.04} & \textbf{16.6} & \textbf{32.35} \\
\cdashline{1-4}
    \multicolumn{4}{l}{\textit{StructCoder} on validation} \\
    \hspace{1ex}replicate & 37.30 & 18.2  & 40.42 \\
    \hspace{1ex}\hspace{1ex}+ augmentation & \textbf{37.48} & \textbf{18.7} & \textbf{40.47} \\
\midrule
    \multicolumn{4}{l}{\textit{CodeGPT} on test}\\
    \hspace{1ex}paper & 32.79 & 20.1 & 35.98 \\
    \hspace{1ex}replicate & 32.66 & \textbf{20.1} & 35.89 \\
    \hspace{1ex}\hspace{1ex}+ augmentation & \textbf{33.45} & 19.2 & \textbf{36.47} \\
\cdashline{1-4}
    \multicolumn{4}{l}{\textit{StructCoder} on test}\\
    \hspace{1ex}paper & 40.91 & 22.4 & 44.77 \\
    \hspace{1ex}replicate & \textbf{41.57} & \textbf{22.6} & \textbf{44.61} \\
    \hspace{1ex}\hspace{1ex}+ augmentation & 41.32 & 21.4 & 44.04 \\
\bottomrule
\end{tabular}

%% file: analysis-appendix-code-number.tex
\begin{lstlisting}[basicstyle=\linespread{0.75}\ttfamily\footnotesize,escapechar=@]
// @\textit{test \#131}@
@\textbf{C\# source}\hspace{7ex}@ public ScaleClusterRequest(): base("CS", "@\textbf{2015}@-12-15", "ScaleCluster"
@\hspace{25.5ex}@ , "cs", "openAPI"){UriPattern = "/clusters/[ClusterId]";
@\hspace{25.5ex}@ Method = MethodType.PUT;}
@\textbf{Java reference}@ public ScaleClusterRequest() {super("CS", "@\textbf{2015}@-12-15", "ScaleCluster"
@\hspace{25.5ex}@ , "csk");setUriPattern("/clusters/[ClusterId]");
@\hspace{25.5ex}@ setMethod(MethodType.PUT);}
@\textbf{baseline}\hspace{8.5ex}@ publicscaleClusterRequest() {super("CS", "@\textbf{2018}@-12-15", "ScaleCluster"
@\hspace{25.5ex}@ , "cs");setUriPattern("/clusters/[ClusterId]");
@\hspace{25.5ex}@ setMethod(MethodType.PUT);}
@\textbf{DA model}\hspace{8.5ex}@ public ClusterRequest() {super("CS", "@\textbf{2015}@-12-15", "ScaleCluster"
@\hspace{25.5ex}@ , "cs");setUriPattern("/clusters/[ClusterId]");
@\hspace{25.5ex}@ setMethod(MethodType.PUT);}

// @\textit{test \#436}@
@\textbf{C\# source}\hspace{7ex}@ public void CopyTo(byte[] b, int o){FormatHexByte(b, @\textbf{o + 0, w1}@);
@\hspace{25.5ex}@ FormatHexByte(b, @\textbf{o + 8, w2}@);FormatHexByte(b, @\textbf{o + 16, w3}@);
@\hspace{25.5ex}@ FormatHexByte(b, o + 24, w4);FormatHexByte(b, o + 32, w5);}
@\textbf{Java reference}@ public void copyTo(byte[] b, int o) {formatHexByte(b, @\textbf{o + 0, w1}@);
@\hspace{25.5ex}@ formatHexByte(b, @\textbf{o + 8, w2}@);formatHexByte(b, @\textbf{o + 16, w3}@);
@\hspace{25.5ex}@ formatHexByte(b, o + 24, w4);formatHexByte(b, o + 32, w5);}
@\textbf{baseline}\hspace{8.5ex}@ public void copyTo(byte[] b, int o) {formatHexByte(b, @\textbf{o1}@);
@\hspace{25.5ex}@ formatHexByte(b, @\textbf{o2}@);formatHexByte(b, @\textbf{o2}@);
@\hspace{25.5ex}@ formatHexByte(b, @\textbf{o3}@);formatHexByte(b,o + 24, w4);
@\hspace{25.5ex}@ formatHexByte(b, o + 32, w5);}
@\textbf{DA model}\hspace{8.5ex}@ public void copyTo(int[] b, int o) {formatHexByte(b, @\textbf{o + 0, w1}@);
@\hspace{25.5ex}@ formatHexByte(b, @\textbf{o + 8, w2}@);formatHexByte(b, @\textbf{o + 16, w3}@);
@\hspace{25.5ex}@ formatHexByte(b, o + 24, w4);formatHexByte(b, o + 32, w5);}

// @\textit{test \#716}@
@\textbf{C\# source}\hspace{7ex}@ public override void Decode(byte[] blocks, int blocksOffset, int[] 
@\hspace{25.5ex}@ values, int valuesOffset, int iterations){for (int j = 0; 
@\hspace{25.5ex}@ j < iterations; ++j){var block = blocks[blocksOffset++];
@\hspace{25.5ex}@ values[valuesOffset++] = ((int)((uint)block >> 7)) & 1;
@\hspace{25.5ex}@ values[valuesOffset++] = ((int)((uint)block >> 6)) & 1;
@\hspace{25.5ex}@ values[valuesOffset++] = ((int)((uint)block >> 5)) & 1;
@\hspace{25.5ex}@ values[valuesOffset++] = ((int)((uint)block >> 4)) & 1;
@\hspace{25.5ex}@ values[valuesOffset++] = ((int)((uint)block >> @\textbf{3}@)) & 1;
@\hspace{25.5ex}@ values[valuesOffset++] = ((int)((uint)block >> 2)) & 1;
@\hspace{25.5ex}@ values[valuesOffset++] = ((int)((uint)block >> 1)) & 1;
@\hspace{25.5ex}@ values[valuesOffset++] = block & 1;}}
@\textbf{Java reference}@ public void decode(byte[] blocks, int blocksOffset, int[] 
@\hspace{25.5ex}@ values, int valuesOffset, int iterations) {for (int j = 0; 
@\hspace{25.5ex}@ j < iterations; ++j) {final byte block = blocks[blocksOffset++];
@\hspace{25.5ex}@ values[valuesOffset++] = (block >>> 7) & 1;
@\hspace{25.5ex}@ values[valuesOffset++] = (block >>> 6) & 1;
@\hspace{25.5ex}@ values[valuesOffset++] = (block >>> 5) & 1;
@\hspace{25.5ex}@ values[valuesOffset++] = (block >>> 4) & 1;
@\hspace{25.5ex}@ values[valuesOffset++] = (block >>> @\textbf{3}@) & 1;
@\hspace{25.5ex}@ values[valuesOffset++] = (block >>> 2) & 1;
@\hspace{25.5ex}@ values[valuesOffset++] = (block >>> 1) & 1;
@\hspace{25.5ex}@ values[valuesOffset++] = block & 1;}}
@\textbf{baseline}\hspace{8.5ex}@ public void decode(byte[] blocks, int blocksOffset, int[] 
@\hspace{25.5ex}@ values, int valuesOffset, int iterations) {for (int j = 0; 
@\hspace{25.5ex}@ j < iterations; ++j) {final byte block = blocks[blocksOffset++];
@\hspace{25.5ex}@ values[valuesOffset++] = (block >>> 7) & 1;
@\hspace{25.5ex}@ values[valuesOffset++] = (block >>> 6) & 1;
@\hspace{25.5ex}@ values[valuesOffset++] = (block >>> 5) & 1;
@\hspace{25.5ex}@ values[valuesOffset++] = (block >>> 4) & 1;
@\hspace{25.5ex}@ values[valuesOffset++] = (block >>> @\textbf{4}@) & 1;
@\hspace{25.5ex}@ values[valuesOffset++] = (block >>> 2) & 1;
@\hspace{25.5ex}@ values[valuesOffset++] = (block >>> 1) & 1;
@\hspace{25.5ex}@ values[valuesOffset++] = block & 1;}}
@\textbf{DA model}\hspace{8.5ex}@ public void decode(byte[] blocks, int blocksOffset, int[] 
@\hspace{25.5ex}@ values, int valuesOffset, int iterations) {for (int j = 0; 
@\hspace{25.5ex}@ j < iterations; ++j) {final byte block = blocks[blocksOffset++];
@\hspace{25.5ex}@ values[valuesOffset++] = (block >>> 7) & 1;
@\hspace{25.5ex}@ values[valuesOffset++] = (block >>> 6) & 1;
@\hspace{25.5ex}@ values[valuesOffset++] = (block >>> 5) & 1;
@\hspace{25.5ex}@ values[valuesOffset++] = (block >>> 4) & 1;
@\hspace{25.5ex}@ values[valuesOffset++] = (block >>> @\textbf{3}@) & 1;
@\hspace{25.5ex}@ values[valuesOffset++] = (block >>> 2) & 1;
@\hspace{25.5ex}@ values[valuesOffset++] = (block >>> 1) & 1;
@\hspace{25.5ex}@ values[valuesOffset++] = block & 1;}}
\end{lstlisting}

%% file: acl_latex.bbl
\begin{thebibliography}{25}
\expandafter\ifx\csname natexlab\endcsname\relax\def\natexlab#1{#1}\fi

\bibitem[{Ahmad et~al.(2021)Ahmad, Chakraborty, Ray, and
  Chang}]{ahmad-etal-2021-unified}
Wasi Ahmad, Saikat Chakraborty, Baishakhi Ray, and Kai-Wei Chang. 2021.
\newblock \href {https://doi.org/10.18653/v1/2021.naacl-main.211} {Unified
  pre-training for program understanding and generation}.
\newblock In \emph{Proceedings of the 2021 Conference of the North American
  Chapter of the Association for Computational Linguistics: Human Language
  Technologies}.

\bibitem[{Ahmad et~al.(2022)Ahmad, Chakraborty, Ray, and
  Chang}]{ahmad-etal-2022-summarize}
Wasi~Uddin Ahmad, Saikat Chakraborty, Baishakhi Ray, and Kai-Wei Chang. 2022.
\newblock \href {https://arxiv.org/abs/2205.11116v1} {Summarize and generate to
  back-translate: Unsupervised translation of programming languages}.
\newblock \emph{arXiv preprint}, abs/2205.11116v1.

\bibitem[{Chen et~al.(2021)Chen, Tworek, Jun, Yuan, Ponde, Kaplan, Edwards,
  Burda, Joseph, Brockman, Ray, Puri, Krueger, Petrov, Khlaaf, Sastry, Mishkin,
  Chan, Gray, ..., and Zaremba}]{chen-etal-2021-evaluating-large}
Mark Chen, Jerry Tworek, Heewoo Jun, Qiming Yuan, Henrique Ponde, Jared Kaplan,
  Harrison Edwards, Yura Burda, Nicholas Joseph, Greg Brockman, Alex Ray, Raul
  Puri, Gretchen Krueger, Michael Petrov, Heidy Khlaaf, Girish Sastry, Pamela
  Mishkin, Brooke Chan, Scott Gray, ..., and Wojciech Zaremba. 2021.
\newblock \href {https://arxiv.org/abs/2107.03374} {Evaluating large language
  models trained on code}.
\newblock \emph{arXiv preprint}, abs/2107.03374v2.

\bibitem[{Christopoulou et~al.(2022)Christopoulou, Lampouras, Gritta, Zhang,
  Guo, Li, Zhang, Xiao, Shen, Li, Yu, Yan, Zhou, Wang, Ma, Iacobacci, Wang,
  Liang, Wei, ..., and Liu}]{christopoulou-etal-2022-pangu}
Fenia Christopoulou, Gerasimos Lampouras, Milan Gritta, Guchun Zhang, Yinpeng
  Guo, Zhong-Yi Li, Qi~Zhang, Meng Xiao, Bo~Shen, Lin Li, Hao Yu, Li~Yan,
  Pingyi Zhou, Xin Wang, Yu~Ma, Ignacio Iacobacci, Yasheng Wang, Guangtai
  Liang, Jia Wei, ..., and Qun Liu. 2022.
\newblock \href {https://arxiv.org/abs/2207.11280} {Pangu-coder: Program
  synthesis with function-level language modeling}.
\newblock \emph{arXiv preprint}, abs/2207.11280v1.

\bibitem[{Clark et~al.(2020)Clark, Luong, Le, and
  Manning}]{clark-etal-2020-electra}
Kevin Clark, Minh-Thang Luong, Quoc~V. Le, and Christopher~D. Manning. 2020.
\newblock \href {https://openreview.net/forum?id=r1xMH1BtvB} {Electra:
  Pre-training text encoders as discriminators rather than generators}.
\newblock In \emph{International Conference on Learning Representations}.

\bibitem[{Clement et~al.(2020)Clement, Drain, Timcheck, Svyatkovskiy, and
  Sundaresan}]{clement-etal-2020-pymt5}
Colin Clement, Dawn Drain, Jonathan Timcheck, Alexey Svyatkovskiy, and Neel
  Sundaresan. 2020.
\newblock \href {https://doi.org/10.18653/v1/2020.emnlp-main.728} {{P}y{MT}5:
  multi-mode translation of natural language and python code with
  transformers}.
\newblock In \emph{Proceedings of the 2020 Conference on Empirical Methods in
  Natural Language Processing (EMNLP)}.

\bibitem[{Currey et~al.(2017)Currey, Miceli~Barone, and
  Heafield}]{currey-etal-2017-copied}
Anna Currey, Antonio~Valerio Miceli~Barone, and Kenneth Heafield. 2017.
\newblock \href {https://doi.org/10.18653/v1/W17-4715} {Copied monolingual data
  improves low-resource neural machine translation}.
\newblock In \emph{Proceedings of the Second Conference on Machine
  Translation}.

\bibitem[{Devlin et~al.(2019)Devlin, Chang, Lee, and
  Toutanova}]{devlin-etal-2019-bert}
Jacob Devlin, Ming-Wei Chang, Kenton Lee, and Kristina Toutanova. 2019.
\newblock \href {https://doi.org/10.18653/v1/N19-1423} {{BERT}: Pre-training of
  deep bidirectional transformers for language understanding}.
\newblock In \emph{Proceedings of the 2019 Conference of the North {A}merican
  Chapter of the Association for Computational Linguistics: Human Language
  Technologies, Volume 1 (Long and Short Papers)}.

\bibitem[{Feng et~al.(2020)Feng, Guo, Tang, Duan, Feng, Gong, Shou, Qin, Liu,
  Jiang, and Zhou}]{feng-etal-2020-codebert}
Zhangyin Feng, Daya Guo, Duyu Tang, Nan Duan, Xiaocheng Feng, Ming Gong, Linjun
  Shou, Bing Qin, Ting Liu, Daxin Jiang, and Ming Zhou. 2020.
\newblock \href {https://doi.org/10.18653/v1/2020.findings-emnlp.139}
  {{C}ode{BERT}: A pre-trained model for programming and natural languages}.
\newblock In \emph{Findings of the Association for Computational Linguistics:
  EMNLP 2020}.

\bibitem[{Guo et~al.(2022)Guo, Lu, Duan, Wang, Zhou, and
  Yin}]{guo-etal-2022-unixcoder}
Daya Guo, Shuai Lu, Nan Duan, Yanlin Wang, Ming Zhou, and Jian Yin. 2022.
\newblock \href {https://doi.org/10.18653/v1/2022.acl-long.499} {{U}ni{X}coder:
  Unified cross-modal pre-training for code representation}.
\newblock In \emph{Proceedings of the 60th Annual Meeting of the Association
  for Computational Linguistics (Volume 1: Long Papers)}.

\bibitem[{Guo et~al.(2021)Guo, Ren, Lu, Feng, Tang, Liu, Zhou, Duan,
  Svyatkovskiy, Fu, Tufano, Deng, Clement, Drain, Sundaresan, Yin, Jiang, and
  Zhou}]{guo-etal-2021-graphcodebert}
Daya Guo, Shuo Ren, Shuai Lu, Zhangyin Feng, Duyu Tang, Shujie Liu, Long Zhou,
  Nan Duan, Alexey Svyatkovskiy, Shengyu Fu, Michele Tufano, Shao~Kun Deng,
  Colin Clement, Dawn Drain, Neel Sundaresan, Jian Yin, Daxin Jiang, and Ming
  Zhou. 2021.
\newblock \href {https://openreview.net/forum?id=jLoC4ez43PZ} {{GraphCodeBERT}:
  Pre-training code representations with data flow}.
\newblock In \emph{International Conference on Learning Representations}.

\bibitem[{Husain et~al.(2019)Husain, Wu, Gazit, Allamanis, and
  Brockschmidt}]{husain-etal-2019-codesearchnet}
Hamel Husain, Hongqi Wu, Tiferet Gazit, Miltiadis Allamanis, and Marc
  Brockschmidt. 2019.
\newblock \href {https://arxiv.org/abs/1909.09436} {{CodeSearchNet} challenge:
  Evaluating the state of semantic code search}.
\newblock \emph{arXiv preprint}, abs/1909.09436v3.

\bibitem[{Iyer et~al.(2018)Iyer, Konstas, Cheung, and
  Zettlemoyer}]{iyer-etal-2018-mapping}
Srinivasan Iyer, Ioannis Konstas, Alvin Cheung, and Luke Zettlemoyer. 2018.
\newblock \href {https://doi.org/10.18653/v1/D18-1192} {Mapping language to
  code in programmatic context}.
\newblock In \emph{Proceedings of the 2018 Conference on Empirical Methods in
  Natural Language Processing}.

\bibitem[{Kanade et~al.(2020)Kanade, Maniatis, Balakrishnan, and
  Shi}]{kanade-etal-2020-learning}
Aditya Kanade, Petros Maniatis, Gogul Balakrishnan, and Kensen Shi. 2020.
\newblock \href {https://dl.acm.org/doi/abs/10.5555/3524938.3525412} {Learning
  and evaluating contextual embedding of source code}.
\newblock In \emph{Proceedings of the 37th International Conference on Machine
  Learning}.

\bibitem[{Kocmi et~al.(2022)Kocmi, Bawden, Bojar, Dvorkovich, Federmann,
  Fishel, Gowda, Graham, Grundkiewicz, Haddow, Knowles, Koehn, Monz, Morishita,
  Nagata, Nakazawa, Nov{\'a}k, Popel, and
  Popovi{\'c}}]{kocmi-etal-2022-findings}
Tom Kocmi, Rachel Bawden, Ond{\v{r}}ej Bojar, Anton Dvorkovich, Christian
  Federmann, Mark Fishel, Thamme Gowda, Yvette Graham, Roman Grundkiewicz,
  Barry Haddow, Rebecca Knowles, Philipp Koehn, Christof Monz, Makoto
  Morishita, Masaaki Nagata, Toshiaki Nakazawa, Michal Nov{\'a}k, Martin Popel,
  and Maja Popovi{\'c}. 2022.
\newblock \href {https://aclanthology.org/2022.wmt-1.1} {Findings of the 2022
  conference on machine translation ({WMT}22)}.
\newblock In \emph{Proceedings of the Seventh Conference on Machine Translation
  (WMT)}.

\bibitem[{Liu et~al.(2019)Liu, Ott, Goyal, Du, Joshi, Chen, Levy, Lewis,
  Zettlemoyer, and Stoyanov}]{liu-etal-2019-roberta}
Yinhan Liu, Myle Ott, Naman Goyal, Jingfei Du, Mandar Joshi, Danqi Chen, Omer
  Levy, Mike Lewis, Luke Zettlemoyer, and Veselin Stoyanov. 2019.
\newblock \href {https://arxiv.org/abs/1907.11692} {Roberta: A robustly
  optimized {BERT} pretraining approach}.
\newblock \emph{arXiv preprint}, abs/1907.11692v1.

\bibitem[{Lu et~al.(2021)Lu, Guo, Ren, Huang, Svyatkovskiy, Blanco, Clement,
  Drain, Jiang, Tang, Li, Zhou, Shou, Zhou, Tufano, Gong, Zhou, Duan,
  Sundaresan, ..., and Liu}]{lu-etal-2021-codexglue}
Shuai Lu, Daya Guo, Shuo Ren, Junjie Huang, Alexey Svyatkovskiy, Ambrosio
  Blanco, Colin Clement, Dawn Drain, Daxin Jiang, Duyu Tang, Ge~Li, Lidong
  Zhou, Linjun Shou, Long Zhou, Michele Tufano, Ming Gong, Ming Zhou, Nan Duan,
  Neel Sundaresan, ..., and Shujie Liu. 2021.
\newblock \href {https://openreview.net/forum?id=6lE4dQXaUcb} {Code{XGLUE}: A
  machine learning benchmark dataset for code understanding and generation}.
\newblock In \emph{Thirty-fifth Conference on Neural Information Processing
  Systems Datasets and Benchmarks Track}.

\bibitem[{Papineni et~al.(2002)Papineni, Roukos, Ward, and
  Zhu}]{papineni-etal-2002-bleu}
Kishore Papineni, Salim Roukos, Todd Ward, and Wei-Jing Zhu. 2002.
\newblock \href {https://doi.org/10.3115/1073083.1073135} {{B}leu: a method for
  automatic evaluation of machine translation}.
\newblock In \emph{Proceedings of the 40th Annual Meeting of the Association
  for Computational Linguistics}.

\bibitem[{Ren et~al.(2020)Ren, Guo, Lu, Zhou, Liu, Tang, Zhou, Blanco, and
  Ma}]{ren-etal-2020-codebleu}
Shuo Ren, Daya Guo, Shuai Lu, Long Zhou, Shujie Liu, Duyu Tang, M.~Zhou,
  Ambrosio Blanco, and Shuai Ma. 2020.
\newblock \href {https://arxiv.org/abs/2009.10297} {{CodeBLEU}: a method for
  automatic evaluation of code synthesis}.
\newblock \emph{arXiv preprint}, abs/2009.10297v2.

\bibitem[{Roziere et~al.(2020)Roziere, Lachaux, Chanussot, and
  Lample}]{roziere-etal-2020-unsupervised}
Baptiste Roziere, Marie-Anne Lachaux, Lowik Chanussot, and Guillaume Lample.
  2020.
\newblock \href
  {https://proceedings.neurips.cc/paper/2020/file/ed23fbf18c2cd35f8c7f8de44f85c08d-Paper.pdf}
  {Unsupervised translation of programming languages}.
\newblock In \emph{Advances in Neural Information Processing Systems}.

\bibitem[{Sennrich et~al.(2016)Sennrich, Haddow, and
  Birch}]{sennrich-etal-2016-improving}
Rico Sennrich, Barry Haddow, and Alexandra Birch. 2016.
\newblock \href {https://doi.org/10.18653/v1/P16-1009} {Improving neural
  machine translation models with monolingual data}.
\newblock In \emph{Proceedings of the 54th Annual Meeting of the Association
  for Computational Linguistics (Volume 1: Long Papers)}.

\bibitem[{Tipirneni et~al.(2022)Tipirneni, Zhu, and
  Reddy}]{tipirneni-etal-2022-structcoder}
Sindhu Tipirneni, Ming Zhu, and Chandan~K. Reddy. 2022.
\newblock \href {https://arxiv.org/abs/2206.05239} {Structcoder:
  Structure-aware transformer for code generation}.
\newblock \emph{arXiv preprint}, abs/2206.05239v1.

\bibitem[{Wang et~al.(2021)Wang, Wang, Joty, and Hoi}]{wang-etal-2021-codet5}
Yue Wang, Weishi Wang, Shafiq Joty, and Steven~C.H. Hoi. 2021.
\newblock \href {https://aclanthology.org/2021.emnlp-main.685} {{C}ode{T}5:
  Identifier-aware unified pre-trained encoder-decoder models for code
  understanding and generation}.
\newblock In \emph{Proceedings of the 2021 Conference on Empirical Methods in
  Natural Language Processing}.

\bibitem[{You and Yim(2010)}]{you-yim-2010-malware}
Ilsun You and Kangbin Yim. 2010.
\newblock \href {https://doi.org/10.1109/BWCCA.2010.85} {Malware obfuscation
  techniques: A brief survey}.
\newblock In \emph{2010 International Conference on Broadband, Wireless
  Computing, Communication and Applications}.

\bibitem[{Yu et~al.(2022)Yu, Wang, and Wang}]{yu-etal-2022-data}
Shiwen Yu, Ting Wang, and Ji~Wang. 2022.
\newblock \href {https://doi.org/https://doi.org/10.1016/j.jss.2022.111304}
  {Data augmentation by program transformation}.
\newblock \emph{Journal of Systems and Software}, 190:111304.

\end{thebibliography}
